\documentclass[sn-mathphys-num]{sn-jnl}

\usepackage{graphicx}%
\usepackage{multirow}%
\usepackage{amsmath,amssymb,amsfonts}%
\usepackage{amsthm}%
\usepackage{mathrsfs}%
\usepackage[title]{appendix}%
\usepackage{xcolor}%
\usepackage{textcomp}%
\usepackage{manyfoot}%
\usepackage{algorithm}%
\usepackage{algpseudocode}%

\newtheorem{definition}{Definition}
\newtheorem{proposition}{Proposition}
\usepackage[final]{changes}
\usepackage{booktabs}

\begin{document}

\title[Article Title]{PretopoMD: Pretopology-based Mixed Data Hierarchical Clustering}


\author*[1]{\fnm{Loup-Noé} \sur{Levy}}\email{loup-noe.levy@energisme.com}
\equalcont{These authors contributed equally to this work.}

\author*[2,3]{\fnm{Guillaume} \sur{Guerard}}\email{guillaume.guerard@devinci.fr}
\equalcont{These authors contributed equally to this work.}

\author[2]{\fnm{Sonia} \sur{Djebali}}\email{sonia.djebali@devinci.fr}

\author[3]{\fnm{Soufian} \sur{Ben Amor}}\email{soufian.ben-amor@uvsq.fr}

\affil*[1]{\orgname{Energisme}, \orgaddress{\street{88 Avenue du Général Leclerc}, \city{Boulogne-Billancourt}, \postcode{92100}, \country{France}}}

\affil[2]{\orgname{Léonard de Vinci Pôle Universitaire, Research Center}, \orgaddress{\street{12 Avenue Léonard de Vinci}, \city{Paris La Défense}, \postcode{92916}, \country{France}}}

\affil[3]{\orgname{LI-PARAD Laboratory EA 7432}, \orgaddress{\street{Versailles University, 55 Avenue de Paris}, \city{Versailles}, \postcode{78035}, \country{France}}}


\abstract{This article presents a novel pretopology-based algorithm designed to address the challenges of clustering mixed data without the need for dimensionality reduction. Leveraging Disjunctive Normal Form, our approach formulates customizable logical rules and adjustable hyperparameters that allow for user-defined hierarchical cluster construction and facilitate tailored solutions for heterogeneous datasets. Through hierarchical dendrogram analysis and comparative clustering metrics, our method demonstrates superior performance by accurately and interpretably delineating clusters directly from raw data, thus preserving data integrity. Empirical findings highlight the algorithm’s robustness in constructing meaningful clusters and reveal its potential in overcoming issues related to clustered data explainability. The novelty of this work lies in its departure from traditional dimensionality reduction techniques and its innovative use of logical rules that enhance both cluster formation and clarity, thereby contributing a significant advancement to the discourse on clustering mixed data.}

\keywords{Mixed Data, Clustering, Machine Learning, Pretopology}



\maketitle

\section{Introduction}

Clustering is a fundamental task in unsupervised machine learning, aimed at grouping similar data points based on inherent similarities in large datasets. While numerous clustering methods have been developed for numerical data, challenges arise when datasets consist of a mixture of numerical, ordinal, and categorical features. Mixed data clustering is critical in various domains, such as the energy sector, biology, medicine, marketing, and economics \cite{abumalloh2024exploring,ahmad2019survey,caruso2021cluster,han2021exploring,mcparland2017clustering,yu2006kernel}. In these fields, datasets often present rich and heterogeneous characteristics, requiring methods that can accommodate different data types without oversimplification.

One of the primary difficulties in clustering mixed data is the diversity in scales, distributions, and similarity measures that characterize different feature types. Traditional clustering techniques, which often rely on a single distance measure or assume data homogeneity, may fail to capture the complex interactions present in heterogeneous datasets. An alternative strategy involves preprocessing the mixed data into a purely numerical format, thereby enabling the use of advanced numerical clustering algorithms. However, this conversion process may lead to significant information loss and an oversimplified representation of the underlying data structure.

The increasing need for robust mixed data clustering methods has spurred significant research. Yet, many existing techniques depend on dimensionality reduction—which can obscure the original data complexity—or lack the flexibility and interpretability required for real-world applications. To overcome these limitations, our work presents a novel clustering algorithm based on the mathematical framework of Pretopology. This approach offers several distinctive advantages over traditional methods.

The innovations and contributions of our paper are as follows:
\begin{itemize}
    \item Direct Handling of Mixed Data: Our algorithm is specifically designed to operate directly on mixed datasets without requiring prior dimensionality reduction. This ensures that the intrinsic complexity of the data is preserved, leading to more authentic and reliable clustering outcomes.
    \item Customizable Cluster Formation via Logical Rules: A key innovation is the incorporation of user-defined logical rules for cluster construction. This feature provides increased flexibility, allowing the clustering process to be tailored to the specific characteristics of the dataset under study.
    \item Hierarchical Flexibility Through Multiple Hyperparameters: Multiple hyperparameters govern the conditions for clustering and division. This flexibility enables the generation of customized hierarchical structures that can reveal underlying patterns at various levels of granularity.
    \item Explainability of the Clustering Process: The algorithm inherently produces a hierarchical dendrogram, offering transparent insights into how clusters are formed and how different hyperparameters influence the segmentation. This level of explainability is particularly valuable in applications where interpretability is essential for validation and decision-making.
\end{itemize}

By addressing the inherent challenges of clustering mixed data and explicitly incorporating innovations that include direct data handling, flexible rule-based construction, customizable hierarchical structuring, and enhanced explainability, our work contributes a robust and interpretable tool to the field of unsupervised machine learning.

The remainder of the paper is organized as follows. Section 2 provides an extensive review of the state-of-the-art in mixed data clustering methods and dimensionality reduction techniques, highlighting their respective strengths and limitations. In Section 3, we introduce our novel pretopology-based clustering algorithm, detailing both its theoretical foundation and its operational mechanics. Section 4 presents experimental results and discussions, demonstrating the effectiveness of our approach across diverse mixed datasets without the need for prior dimensionality reduction.

\section{Literature review}
\subsection{Machine Learning Models}

Mixed data clustering involves grouping datasets with a combination of numerical, categorical, and other types of features. In this section, we offer an overview of prominent open-source methods in mixed data clustering.

\subsubsection*{Partitioning Clustering}
Partitioning clustering aims to divide a dataset into a predefined number of non-overlapping clusters. Two notable methods in this category are:

\begin{itemize}
\item \textbf{K-Prototypes} \cite{huang1998extensions}: An extension of traditional K-Means, K-Prototypes handles both numerical and categorical features. It combines K-Means for numerical features with the K-Modes method \cite{huang1997clustering} for categorical features, utilizing a model-based approach for clustering categorical data.

\item \textbf{Convex K-Means} \cite{modha2003feature}: This algorithm refines centroids with convex hull iteratively, aiming for convergence or a predefined maximum number of iterations.
\end{itemize}

\subsubsection*{Model-Based Clustering}
Model-based clustering leverages statistical models to describe data distribution within clusters, accommodating various data types. Key methods include:

\begin{itemize}
\item{\textbf{MixtComp}} \cite{biernacki2016bigstat}: A statistical approach for clustering mixed data, MixtComp combines model-based clustering and Bayesian approaches. It models mixed data as a mixture of multivariate distributions, utilizing latent variables to capture complex data structures.

\item{\textbf{KAMILA}} (KAy-means for MIxed LArge data) \cite{foss2016semiparametric}: Extending the standard K-means algorithm, KAMILA handles mixed data through a combination of K-means clustering and the Gaussian-multinomial mixture model. It avoids strong parametric assumptions and balances the effects of numerical and categorical data.

\item{\textbf{ClustMD}} (Clustering of Mixed Data) \cite{mcparland2016model}: A model-based clustering method specifically designed for mixed data, ClustMD employs a latent variable model fitted with the Expectation-Maximization algorithm.
\end{itemize}

\subsubsection*{Hierarchical Clustering}
Hierarchical clustering is widely used for grouping similar data points based on similarity or distance measures. Notable hierarchical methods for mixed data include:

\begin{itemize}
\item{\textbf{Phillip and Ottaway}} \cite{philip1983mixed}: This method proposes hierarchical clustering for mixed data based on Gower's similarity measure. It separates categorical and numerical features, computes similarities in each feature space, and combines them to create a similarity matrix, facilitating meaningful clustering of mixed data.

\end{itemize}

\subsection{Deep Learning Models}

Deep learning models—including convolutional neural networks (CNNs), recurrent neural networks (RNNs), and transformers—have demonstrated remarkable capabilities in automating the extraction of high-level features from raw data. They have achieved state-of-the-art performance in areas such as image recognition, natural language processing, and speech recognition by learning complex, hierarchical representations. These models can autonomously capture and refine intricate patterns from large-scale datasets, which not only minimizes the need for extensive manual feature engineering but also leads to considerable successes in diverse domains. Advanced techniques—such as contrastive learning, deep clustering, and methods incorporating self-labeling and pseudo-label refinement—further extend their capabilities in challenging unsupervised or semi-supervised tasks \cite{liu2024rpsc,yu2025review}.

Despite these strengths, several significant drawbacks have led us to decide against the development of deep learning models for the current project. A primary concern is their substantial dependency on large, well-annotated datasets. Many deep learning architectures require a critical mass of labeled data to generalize well; when available data is limited, these models become prone to overfitting. This problem is compounded in advanced methods—such as multi-view techniques and soft contrastive losses—that often demand extra iterative training stages (like pseudo-label generation and refinement) and further exacerbate data requirements.

In addition to issues with data dependency, the computational demands of deep learning models are considerable. Training these models typically requires specialized hardware (such as GPUs or TPUs) along with extended processing time and high memory consumption. This level of computational overhead makes them less feasible in environments where such resources are constrained—a concern that is well documented in recent investigations \cite{iqbal2025end,yu2023broad}. Moreover, these models tend to be very sensitive to hyperparameter choices including architectural design, learning rates, and regularization strategies. This sensitivity not only renders the development and tuning process time-consuming and error-prone but also impedes the efficient deployment of these large models in production settings.

Another notable drawback is the limited interpretability of deep learning models. Their complex, multilayered structures often result in a "black box" effect, making it difficult to understand the basis of their decision-making. This lack of transparency is particularly problematic for applications requiring clear, explainable insights. Although emerging hybrid architectures and automated hyperparameter tuning methods \cite{chenghu2024novel,shi2023adaptive,yaro2024improved} show promise in mitigating some of these issues, they remain largely experimental and have yet to fully address the underlying challenges.

Given these considerations—the extensive data requirements, high computational demands, and limited interpretability—our current project, which operates under restricted data and resource constraints while demanding robust transparency, motivates the use of alternative methods. By prioritizing approaches that can deliver greater efficiency, robustness, and explainability, our strategy is more closely aligned with the project’s operational requirements and long-term goals.

\subsection{Dimensionality Reduction}

Dimensionality reduction techniques prove valuable in applying advanced numerical data clustering methods to datasets with a mix of numerical and categorical features. However, caution must be exercised as these techniques may lead to substantial information loss and compromise the preservation of intrinsic relationships, impacting the accuracy and interpretability of results.

In existing literature, only Factor Analysis of Mixed Data (FAMD) addresses mixed data dimensionality reduction. Nevertheless, we have adapted two numerical data dimensionality reduction techniques for mixed data:

\begin{itemize}
\item \textbf{Factor Analysis of Mixed Data (FAMD)} \cite{escofier1979traitement}: An extension of Principal Component Analysis (PCA) for continuous variables and Multiple Correspondence Analysis (MCA) for categorical variables, providing a low-dimensional representation of mixed data.
\item \textbf{Uniform Manifold Approximation and Projection (UMAP)}: Originally designed for numerical data, we applied UMAP to mixed data by employing the Huang distance metric, suitable for mixed data. UMAP focuses on preserving local data structure, making it efficient for large-scale datasets.
\item \textbf{Pairwise Controlled Manifold Approximation (PaCMAP)} \cite{wang2021understanding}: Initially designed for numerical data, PaCMAP aims to preserve both local and global structures. We adapted PaCMAP for clustering mixed data by utilizing FAMD instead of PCA in the initialization phase.
\end{itemize}

\subsection{Mixed data cluster evaluation indices}

Key measures for data clustering, categorized into external and internal indices, facilitate the evaluation of clustering algorithms on mixed data. For this paper, three internal measures are employed after FAMD reduction due to the absence of a clustering evaluation method for mixed data. Since clustering is unsupervised learning, the focus is on internal indices independent of ground truths.

\textbf{Calinski-Harabasz Index \cite{calinski1974dendrite}.} Also known as the Variance Ratio Criterion, this index calculates the ratio of between-cluster dispersion to within-cluster dispersion. A higher Calinski-Harabasz score suggests better-defined clusters.

\textbf{Silhouette Coefficient \cite{rousseeuw1987silhouettes}.} This index evaluates the compactness and separation of the clusters. A higher Silhouette Coefficient indicates well-defined clusters. Since the Silhouette coefficient computes pairwise distances, we use it after FAMD reduction or with the Gower distance.

\textbf{Davies Bouldin Index \cite{davies1979cluster}.} This index is the average of the maximum ratio between the distance of a point from the center of its group and the distance between two centers of groups.

It is crucial to recognize that in mixed data clustering, cluster evaluation indices have limitations, including information loss from dimensionality reduction, incompatibility with non-Euclidean spaces, and sensitivity to data characteristics like outliers and skewness. Further details on additional external and internal measures are available in the GitHub repository\footnote{\url{https://github.com/ClementCornet/Benchmark-Mixed-Clustering}}.

\subsection{Summary}

Find in Table \ref{tab:clustering_methods} the pros and cons of the presented algorithms where $\sharp k?$ refers to the hyperparameters $k$ to indicate the number of output clusters.

\begin{table}[ht]
\centering
\begin{tabular}{@{}l l c p{3.5cm} p{3.5cm}@{}}
\toprule
\textbf{Method} & \textbf{Type} & \textbf{$\sharp k?$} & \textbf{Key Strengths} & \textbf{Key Limitations} \\ \midrule
K-Prototypes & Partitional & Yes & Combines Euclidean (numerical) and matching (categorical) distances in a unified framework. & Simplistic categorical distance measure; requires careful tuning of  $ k $  and balancing hyperparameter $\gamma$. \\[0.3em]
Convex K-Means & Partitional & Yes & Refines centroid positions via convex hull concepts; produces more coherent clusters with improved convergence. & Additional convex-combination steps add computational overhead; still requires a predefined cluster count. \\[0.3em]
MixtComp & Model-Based & Yes & Provides a rigorous probabilistic framework; models mixed data with latent variables for rich interpretability. & Computationally intensive; potential convergence issues due to Bayesian estimation. \\[0.3em]
Kamila & Model-Based & Yes & Balances numerical and categorical data using Gaussian–multinomial models; scalable for large datasets. & Sensitive to initialization and parameter settings; requires pre-specification of the number of clusters. \\[0.3em]
ClustMD & Model-Based & Yes & Comprehensive modelling that handles heterogeneity and missing data; models statistical properties of both data types. & Higher computation time; convergence can be challenging for some datasets. \\[0.3em]
Phillip \& Ottaway & Hierarchical & No & Produces an interpretable dendrogram revealing multi-level nested cluster structures without predefining $k$. & Sensitive to merging criteria; may not scale well for very large datasets. \\[0.3em]
PretopoMD & Hierarchical & No & Directly handles mixed data without dimensionality reduction; offers customizable logical rules and multiple hyperparameters; highly explainable via a dendrogram. & Computationally intensive due to iterative pseudoclosure and adjacency matrix construction; performance sensitive to hyperparameter settings. \\[0.3em]
\bottomrule
\end{tabular}
\caption{Summary of mixed data clustering methods.}
\label{tab:clustering_methods}
\end{table}

\section{Theory of Pretopology}
\label{sec:Pretopology}

In this section, we delve into the key concepts and definitions of pretopology as outlined in the article, including pretopological space and pseudo-closure. Although we won't extensively cover the origins of pretopology, it's crucial to grasp that the concept of a pretopological space arises by relaxing the assumptions of topological spaces, enabling the modeling of discrete structures \cite{Auray2009Pretopology}.

\subsection{Pretopological space}

\begin{definition} A pseudoclosure function, denoted as $a : \wp(U) \to \wp(U)$ on a set $U$, is defined by:
\begin{itemize}
    \item $a(\emptyset) = \emptyset$
    \item $\forall A \mid A \subseteq U : A \subseteq a(A)$
\end{itemize}
where $\wp(U)$ is the power set of U.
\end{definition}

\begin{figure}
\centering
\includegraphics[width=0.5\columnwidth]{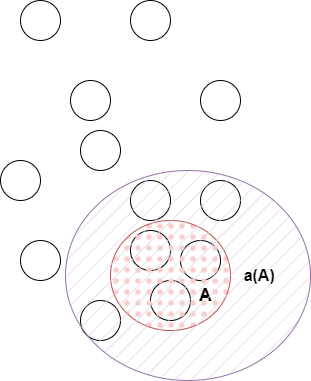}
\caption{Example of a pseudoclosure function.}
\label{fig:pseudoclosure}
\end{figure}

This function establishes a relation between any set of elements and a larger set, forming the basis of a pretopological space, and by the way, a hierarchy. An illustrative example of a pseudoclosure function is depicted in Figure \ref{fig:pseudoclosure}.

\begin{definition} A tuple $(U, a(.))$, where $U$ is a set of elements and $a(.)$ is a pseudoclosure function on $U$, constitutes a pretopological space.
\end{definition}

The most general pretopological space is determined by the above definition. Introducing additional conditions yields more specific pretopological spaces:

\begin{definition} \textbf{Isotony}. If $\forall$ $A, B$ $\vert$ $A \subseteq U$, $B \subseteq U$ : $A \subseteq B$ $\implies$ $a(A) \subseteq a(B)$, then we get a pretopological space of type $V$.
\end{definition}

\begin{definition} \textbf{Union Preservation}. If $\forall$ $A, B$ $\vert$ $A \subseteq U$, $B \subseteq U$ : $a(A \cup B) = a(A) \cup a(B)$, then we get a pretopological space of type $V_D$.
\end{definition}

\begin{definition} \textbf{Union Distribution}. If $\forall$ $A$ $\vert$ $A \subseteq U$ : $a(A) = \bigcup_{x \in A} a(x)$ then we get a pretopological space of type $V_S$.
\end{definition}

In pretopology, the concept of closure aligns with its definition in topology \cite{le2007classification}: 

\begin{definition} A part $F$ of $U$ will be a closure of $U$ if and only if $a(F) = F$
\end{definition}

\begin{proposition} \textbf{Intersection of Closures}. In a pretopological space of type $V$, the intersection of closures is a closure.
\end{proposition}

\begin{proposition} \textbf{Existence of Closure and Opening}. In a pretopological $V-type$ space, the closure and opening of any part of U still exists.
\end{proposition}

\begin{proposition} \textbf{Smallest Closure}. In a pretopological space of type $V$, the closure of a part $A$ of $U$ is the smallest closure containing $A$. Denoted F(A).
\end{proposition}

\begin{proposition} \textbf{Every Set Has a Closure}. In a pretopological space of type $V$, every set has a closure.
\end{proposition}

\begin{figure}
\centering
\includegraphics[width=0.5\columnwidth]{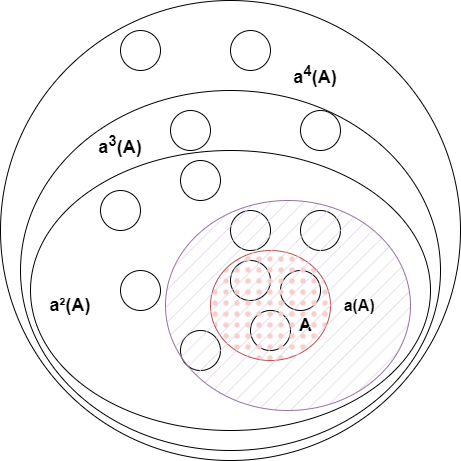}
\caption{Closure of set $A$, $a^4 (A)=F(A)$.}
\label{fig:closure}
\end{figure}

The closure in a pretopological space of type $V$ can be obtained by iteratively applying the pseudoclosure operator to the set and its subsequent images until expansion ceases. Figure \ref{fig:closure} illustrates an example of this process.

\begin{figure}[ht]
  \centering
  \includegraphics[width=0.8\columnwidth]{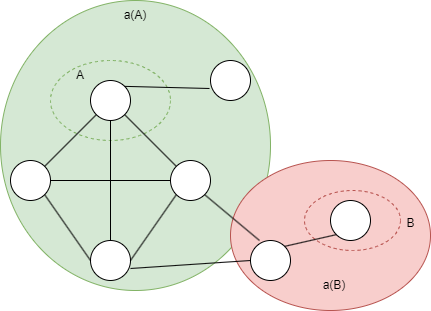}
  \caption{Pseudoclosure function on a graph.}
  \label{fig:pseudoclosure_graph}
\end{figure}

If a pretopological space is of type $V_D$, and for every $A \subseteq U: a(A) = a(a(A))$, the pseudoclosure function is termed idempotent \cite{laborde2019Pretopology}. In such cases, a topology is obtained. Notably, in a finite space, $V_S = V_D$ \cite{belmandt1993manuel}. Additionally, in pretopological spaces of type $V_D$, the pseudoclosure of a set is entirely determined by the pseudoclosures of its singletons. When the space is finite, the pseudoclosure can be represented as a graph, establishing pretopology as a generalization of graph theory \cite{laborde2019Pretopology}. Figure \ref{fig:pseudoclosure_graph} shows the relation between the two.

An alternative characterization involves the concepts of prefilter and neighborhood. To comprehend this, let's introduce the following definitions:

\begin{definition} We say that a set $\mathcal{F}$ of $\wp(\wp(U))$ is a prefilter over $U$, if:
\begin{equation}
\forall F \in \mathcal{F},  \forall H \in \wp(U), F \subset H \Longrightarrow H \in \mathcal{F}
\end{equation}

\end{definition}

\begin{definition} $\mathcal{F}$ of $\wp(\wp(U))$ is a filter over $U$, if it is a prefilter stable under finite intersection, i.e.
\begin{equation}
\forall F \in \mathcal{F}, \forall G \in \mathcal{F}, F \cap G \in \mathcal{F}
\end{equation}
\end{definition}

\begin{figure*}[ht]
  \centering
  \includegraphics[width=\textwidth]{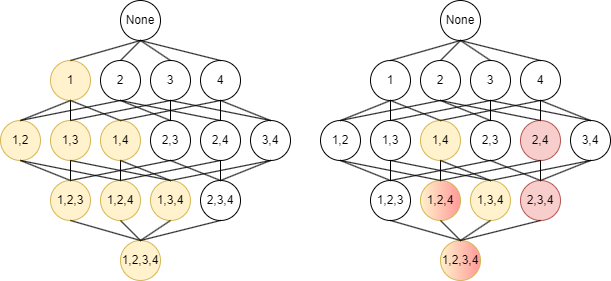}
  \caption{Filters vs Prefilters.}
  \label{fig:filtervsprefilter}
\end{figure*}

In other words, and restricting ourselves to a finite space, a filter is the family of all
supersets of a set $\mathcal{B}$, while a prefilter is the family of supersets of every member $B_i$ of a family of sets $\mathcal{B}$. The family of sets $\mathcal{B}$ is called the basis of the prefilter. 
We can see in Figure \ref{fig:filtervsprefilter} an example of a filter and a prefilter with basis $B$ = {{1, 4}, {2, 4}}.

If we have a set $U$ and, for every $x \in U$, there exists a prefilter $V(x)$ such that every member of $V(x)$ contains the element $x$, we can define a pseudoclosure function as follows:

\begin{equation}
\forall A \subseteq U, a(A) = \{x \in U \vert \forall V \in V (x), V \cap A, \emptyset\}
\end{equation}

\begin{figure*}[ht]
  \centering
  \includegraphics[width=\textwidth]{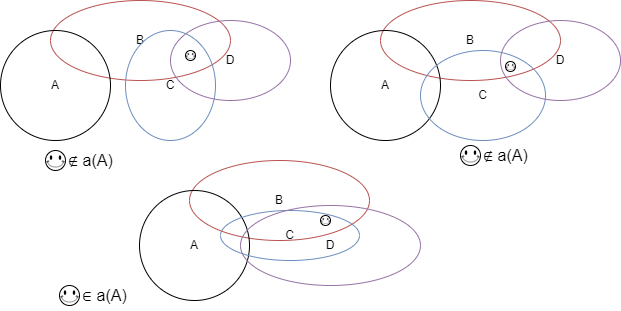}
  \caption{Neighborhood definition of a pretopology.}
  \label{fig:neighborhood}
\end{figure*}

Here, $V(x)$ is termed the family of neighborhoods of $x$, and each set in the family is referred to as a neighborhood of $x$ Figure \ref{fig:neighborhood} shows a graphical representation of this definition of the pseudoclosure.

Conversely, if we have a pseudoclosure function $a(.)$ in a pretopological space of type $V$, the family of sets given by:
\begin{equation}
V (x) = \{V \subset U \vert x \in i(V )\}
\end{equation}
where $i(A) = a(A^c)^c$, is a prefilter.

A proposition asserts that no two families of prefilters or pseudoclosure functions define the same pretopological space, demonstrating the interchangeability of these characterizations \cite{belmandt1993manuel}.


\subsection{Framework}


We introduce a framework for formalizing a pretopological space, building upon the work of Julio Laborde \cite{laborde2019Pretopology}. The pretopological space is defined by a tuple $(G, \Theta, DNF(.))$, where:

\begin{itemize}
\item $G = {G_1 (V_1, E_1 ), G_2 (V_2, E_2 ), ..., G_n (V_n, E_n )}$ represents a collection of $n$ weighted directed graphs.
\item $\Theta = {\theta_1, \theta_2, ..., \theta_n}$ is a set of $n$ thresholds, each associated with a specific graph.
\item $DNF (.) : (\wp(U), U) \to {True, False}$ is a boolean function defined as a positive disjunctive normal form involving the $n$ boolean functions $V_1 (A, x), ..., V_n (A, x)$, each associated with a graph. The truth value depends on the set $A$ and the element $x$.
\end{itemize}

To determine if an element $x \in U$ belongs to the pseudoclosure of a set $A$, the following steps are followed:

\begin{itemize}
\item \textbf{Boolean Function Evaluation}. For each $V_i (A, x)$, $V_i (A, x) = True$ if and only if $\sum_{e_{xy} \in G_i, y \in A} w(e_{xy}) \ge \theta_i$, where $e_{xy}$ denotes the edge from $x$ to $y$, and $w(e)$ represents the weight of the edge $e$.
\item \textbf{Pseudoclosure Membership}. The element $x \in U$ belongs to the pseudoclosure of $A$ if and only if the $DNF(.)$ evaluates to True.
\end{itemize}

In essence, this process checks whether the sum of the edge weights connecting the element $x$ to the elements within $A$ is greater than the threshold associated with the graph in each graph. If this condition is met, the boolean variable corresponding to that graph takes the value True; otherwise, it takes the value False. If $DNF(.)$ evaluates to True given the values of the boolean functions $V_i (A, x)$, then the element belongs to the pseudoclosure. 

\subsection{PretopoMD Algorithm}

This section outlines the Python library's algorithms designed for constructing closures and building hierarchical clustering of mixed data. 

The algorithm \ref{listing:appendix:pretopology:structuralAnalysis}, provided as pseudocode in Algorithm 1, is organized into four stages:
\begin{itemize}
\item \textbf{Identify Seeds}. Identify a family of elementary subsets, referred to as seeds.
\item \textbf{Construct Closures}. Construct closures of seeds through iterative application of the pseudoclosure function.
\item \textbf{Create Adjacency Matrix}. Create the adjacency matrix representing relationships between all recognized subsets, including intermediate ones.
\item \textbf{Determine Quasi-Hierarchy}. Determine the quasi-hierarchy by applying the corresponding algorithm to the adjacency matrix.
\end{itemize}

\begin{algorithm}
\caption{\textbf{QuasistructuralAnalysis}: Algorithm for building a quasi-hierarchy from pretopological space.
}
\begin{algorithmic}
\Require $((U, a(.)), d, seed\_Func(.), th_{qh})$
\Ensure $Sets_{qh}, Adj_{qh}$
\State $seed\_List \leftarrow Set\_Seeds((U, a), d, seed\_Func)$
\State $Sets_{ipc} \leftarrow Iterative\_Pseudoclosure((U, a), seed\_List)$
\State $Atr \leftarrow Attraction\_Matrix(Sets_{ipc})$
\State $Sets_{qh}, Adj_{qh} \leftarrow QuasiHierarchy(Sets_{ipc}, Atr, th_{qh})$
\end{algorithmic}
\label{listing:appendix:pretopology:structuralAnalysis}
\end{algorithm}

At this point, we need to define three hyperparameters:
\begin{itemize}
\item $ seed\_Func(.)$: Determines a set of nearby elements for a given element, constituting a seed.
\item $d$: Specifies the size of the seeds.
\item $th_qh$: Threshold above which two sets are considered related in the hierarchy. This hyperparameter is required by the \textit{Quasi\_Hierarchy} algorithm to establish the quasi-hierarchy.
\end{itemize}

We will now discuss each stage of the algorithm in detail.

\subsubsection*{Computation of a Family of Elementary Sets or Seeds}
\label{paragraph:appendix:codeExamples:algos:pretopology:algorithmConception:algoFamilyElementarySeed}

The objective is to identify elementary subsets of size $d$, commonly known as seeds, utilizing the $seed_Func(.)$ function. This function is tasked with locating the necessary $d$ neighbors. The process involves iterating over all the points within the set $U$, associated with the pretopological space $p$. The pseudocode for the resultant algorithm, denoted as $Elem_Quasiclosures$, is presented in Algorithm \ref{listing:appendix:pretopology:elementaryQuasiclosures}.

\begin{algorithm}
\caption{\textbf{Set\_Seeds}: Construction of the seeds of size $d$ by applying the function $seed\_Func(.)$ on all the elements of the set $U$.
}
\begin{algorithmic}
\Require $((U, a(.)), d, seed\_Func(.))$
\Ensure $seed\_List$
\State $seed\_List \leftarrow list()$
\ForAll{$x \in U$}
\State $seed \leftarrow seed\_Func(x, d)$
\State $seed\_List.append(seed)$
\EndFor
\end{algorithmic}
\label{listing:appendix:pretopology:elementaryQuasiclosures}
\end{algorithm}

\begin{algorithm}
\caption{\textbf{FindNeighbors}: Determine the $d$ neighbors of $firstNode$ using the $seedFunc(.)$ function.
}
\begin{algorithmic}
\Require $(first\_Node, d, seed\_Func(.))$
\Ensure $path$
\State $path \leftarrow list()$
\State $last\_Treated\_Node \leftarrow first\_Node$
\ForAll{$ i \in range(d)$}
\State $new\_Node \leftarrow seed\_Func(last\_Treated\_Node)$
\State $path.append(new\_Node)$
\State $last\_Treated\_Node \leftarrow new\_Node$
\EndFor
\end{algorithmic}
\label{listing:appendix:pretopology:findNeighbors}
\end{algorithm}

Algorithm \ref{listing:appendix:pretopology:elementaryQuasiclosures} utilizes the function $Find_Neighbors$, for which the pseudocode is outlined in Algorithm \ref{listing:appendix:pretopology:findNeighbors}. This function accepts an element from $U$, the desired number of neighbors $d$, and the function that determines the nearest neighbors, denoted as $seed\_Func(.)$. The $seed\_Func(.)$ function typically assumes one of the following two forms:
 
\begin{itemize}
\item This function identifies the nearest nodes to an element. It is employed when a distance can be calculated, such as when the studied relations are quantifiable.
\item This function randomly navigates through neighboring nodes. Its usage is preferred when the relations are not quantifiable, for instance, when values describe categories.
\end{itemize}

\subsubsection*{Creation of Subsets through Iterative Pseudoclosure Applications}
\label{paragraph:appendix:codeExamples:algos:pretopology:algorithmConception:subsetCreation}

The $seed\_Func(.)$ function generates the subsets that will be organized by the quasi-hierarchy algorithm, using the seed list $seed\_List$ computed earlier by \textit{QuasistructuralAnalysis}. For each seed in $seed_List$, the membership function is applied iteratively until the pseudoclosure no longer produces larger sets.

The resulting subsets are stored in a list of sets named $QF_{tmp}$ which organizes the subsets based on the number of elements they contain. The subsets of size $s$ are positioned in the $s$-th slot of $QF_{tmp}$. As the pseudoclosure function $a(.)$ only yields a set that is larger or equal in size, applying the pseudoclosure function to sets in ascending order of size ensures that all elements are processed once and only once.

The list $Sets_{ipc}$ is then constructed from the lists in $QF_{tmp}$ and returned. The corresponding pseudocode is provided in Algorithm \ref{listing:appendix:pretopology:elementaryClosedSubsets}.

\begin{algorithm}
\caption{\textbf{Iterative\_Pseudoclusure}: Calculation of subsets by iterative application of the pseudo-closure function.}
\begin{algorithmic}
\Require $((U, a(.)), seed\_List)$
\Ensure $Sets_{ipc}$
\State $QF_{tmp}$ a list of $Size(U)$ of empty sets
\ForAll{$seed \in seed\_List$}
\State $QF_{tmp}[Size(seed)].append(seed)$
\EndFor
\ForAll{$i \in range(1, Size(U) + 1)$}
\ForAll{$s \in QF_{tmp}[i]$}
\State $a_{s} \leftarrow a(s)$
\If{$a_{s}$ \textbf{not in} lists of $QF_{tmp}$}
\State $QF_{tmp}[Size(a_{s})].append(a_{s})$
\EndIf
\EndFor
\EndFor
\State $Sets_{ipc} \leftarrow list()$
\ForAll{$i \in range(Size(QF_{tmp}))$}
\State $Sets_{ipc}.extend(QF_{tmp}[i])$
\EndFor
\end{algorithmic}
\label{listing:appendix:pretopology:elementaryClosedSubsets}
\end{algorithm}

\subsubsection*{Creation of the Attraction Matrix}
\label{paragraph:appendix:codeExamples:algos:pretopology:algorithmConception:attractionmatrix}

The iterative application of a pseudoclosure to two seeds can generate distinct sets that have non-empty intersections. Traditional hierarchies of sets typically handle sets that either have no intersection or are contained within one another (i.e., subsets and supersets). Therefore, a different type of relationship needs to be defined, referred to as a quasi-hierarchy.

To establish this quasi-hierarchy, an attraction matrix is constructed first, as outlined in Algorithm \ref{listing:appendix:pretopology:attractionmatrix}. This matrix represents the "attraction" that sets exhibit for each other. Here, the term "attraction" denotes a non-symmetrical relationship between two intersecting sets, determined by the sizes of the sets and the size of their intersection. The underlying principles include:

\begin{itemize}
\item Two subsets should only be attracted to each other if their intersection is non-empty (i.e., $A \cap B \neq \emptyset$).
\item The larger the cardinality of the intersection $A \cap B$ relative to that of $A$, the stronger the attraction between $A$ and $B$,
\item The larger the cardinality of the subset $B$ relative to that of $A$, the less critical it is for $A \cap B$ to be large for the relation between $A$ and $B$ to be strong. In other words, a very large set will attract smaller sets even if their intersection is not very large.
\end{itemize}

\begin{algorithm}
\caption{\textbf{Attraction\_Matrix}: Construction of the attraction matrix for the quasihierarchy.
}
\begin{algorithmic}
\Require $(Sets_{ipc})$
\Ensure $Atr$
\State $Atr \leftarrow Squared\_Matrix\_Zeros(size(Sets_{ipc}))$
\ForAll{$A, B \in Sets_{ipc}$}
\State $A\_has\_B \leftarrow Size(A \cap B)/Size(B)$
\State $B\_has\_A \leftarrow Size(A \cap B) / Size(A)$
\State $A\_bigger\_B \leftarrow Size(A) / Size(B)$
\State $B\_bigger\_A \leftarrow Size(B) / Size(A)$
\State $Atr[B\_index, A\_index] = B\_bigger\_A * B\_has\_A$
\State $Atr[A\_index, B\_index] = A\_bigger\_B * A\_has\_B$
\EndFor
\end{algorithmic}
\label{listing:appendix:pretopology:attractionmatrix}
\end{algorithm}

\subsubsection*{Creation of the Quasi-Hierarchy}
\label{paragraph:appendix:codeExamples:algos:pretopology:algorithmConception:subsetHierarchyCreation}

The quasi-hierarchy is characterized by a list of sets and an adjacency matrix. The adjacency matrix is derived from the attraction matrix by assessing whether the attraction values in the attraction matrix surpass the threshold $th_{qh}$. The establishment of the quasi-hierarchy involves applying the following rules to the values of $Atr$:

\begin{itemize}
\item A link (as defined by graph theory) between two subsets is formed in the quasi-hierarchy if their attraction surpasses the threshold $th_{qh}$.
\item Two subsets exhibiting strong mutual attraction (i.e., surpassing the threshold $th_{qh}$) are considered equivalent, and only one of them is retained. If the sets are of equal size, one of them is randomly selected; otherwise, the smaller set is removed.
\item The updated list of sets, along with their adjacency matrix, dictates the structure of the quasi-hierarchy.
\end{itemize}

\begin{algorithm}
\caption{\textbf{Quasi\_Hierarchy}: Ensures QuasiHierarchy}
\begin{algorithmic}
\Require $Sets_{ipc}$, $Atr$, $th_{qh}$
\Ensure $Sets_{qh}$, $Adj_{qh}$
\State $Adj_{qh} \leftarrow Squared\_Matrix\_Zeros(size(Sets_{ipc}))$
\State $Adj_{qh} [Atr > th_{qh}] \leftarrow 1$
\ForAll{$i, j \in Range(size(Sets_{ipc}))$}
    \If{$ Adj_{qh} [i, j] = 1$ \textbf{\&} $Adj_{qh} [j, i] = 1$} 
        \If{$size\_of\_set(i) >= size\_of\_set(j)$}
            \State remove set j from $Adj_{qh}$ and $Sets_{ipc}$
        \Else
            \State remove set i from $Adj_{qh}$ and $Sets_{ipc}$
        \EndIf
    \EndIf
\EndFor
\end{algorithmic}
\label{listing:appendix:pretopology:Quasihierarchy}
\end{algorithm}

\subsection{Hyperparameters}
\label{sub:hyperparameters}

The definition of the pretopological space significantly influences cluster formation. For example, considering all $n$ numeric features together, with their Euclidean distances calculated in one graph of $G$, and separately calculating Hamming distances for all categorical values in another graph of $G$, results in a straightforward pretopological space. The DNF could be a logical "AND" or "OR" combination of the Euclidean and Hamming distances. Alternatively, features can be considered individually, each with its own graph, similarity measure, and threshold, making the DNF more comprehensive and specific.

Thresholds are automatically calculated to adapt to the number of points, the number of close neighbors each point has, and the dispersion in the dataset. Alternatively, they can be set manually. The parameters in the threshold calculation function can be adjusted to obtain either high thresholds, resulting in small clusters with low inner dispersion and a high number of outliers, or lower thresholds, leading to larger clusters with fewer outliers.

The threshold $th_{qh}$ used in constructing the quasi-hierarchy is usually fixed at 0.5.

The DNF function defines the logical rules governing cluster formation. Using a logical AND (i.e., $G_i$ AND $G_j$) creates a more constrained clustering, where clusters exhibit similar values for characteristics $i$ and $j$. Conversely, a logical OR (i.e., $G_i$ OR $G_j$) results in less constrained clusters, where clusters show similar values for either characteristic $i$ or $j$. 

Several functions and values are employed to define the networks specific to each feature and the subsequent pretopological space. They are designed to adapt to the dataset's characteristics, such as the number of elements, the nature of the features, and the dispersion of the values. These functions are also tailored to user needs, such as the approximate size of desired clusters or the extent to which outliers are accepted into clusters. For instance, a higher square length results in fewer clusters, and the higher the threshold, the more outliers are found. Users can adjust these parameters by changing the way they are calculated or by directly modifying the coefficients (or powers).

Here is a list of the hyperparameters:
\begin{itemize} 
     \item $\text{thresholds} = \left(\frac{\text{real\_points}}{\text{size}}\right)^{\text{threshold\_power}}$;
     \item $\text{real\_points} = \text{nb\_elements} - \sum(\text{inverse})$;
     \item $\text{inverse}_i = \frac{\text{closest}_i - 1}{\text{closest}_i}$;
     \item $\text{closest}_i = \text{count(}\ j\ \text{with}\ \text{dist}(i, j) < \text{square\_lgth} \times \text{closest\_coeff})$;
     \item $\text{square\_length} = \sqrt{\frac{\text{area}}{\text{size}}} \times \text{square\_lgth\_coeff}$;
     \item $\text{area} = \text{area\_method}(\text{dm}, \text{df\_data})$;
     \item $\text{dm} = \text{matrix of distances using the distance function}$.
 \end{itemize}

\subsection{Explainability}
\label{XAI}
The proposed clustering method aligns with the principles of eXplainable Artificial Intelligence by offering a clear, customizable, and traceable path from input data to clustered groups. This facilitates intuitive understanding and validation of clusters by domain experts, highlighting the following key features:

\begin{itemize}
    \item \textbf{Customizable Hyperparameters}: Hyperparameters are easily customizable, either through fixed values or adaptive functions that respond to data size, dispersion, and desired outlier count. This tunability empowers experts to exert precise control over the clustering process, enhancing the transparency and interpretability of the formed clusters.
    \item \textbf{Hierarchical Structure}: The clustering method adopts a hierarchical structure, presenting a tree-like representation of data groups. This structure facilitates a step-by-step comprehension of how clusters are formed and their interrelationships, contributing to the transparency of the clustering process.
    \item \textbf{Disjunctive Normal Form}: The DNF explicitly defines the role of each parameter in clustering, ensuring transparency in the logic behind grouping. This formalism contributes to a clear articulation of the reasoning behind the formation of clusters.
    \item \textbf{Adaptation to Mixed Data Types}: The method accommodates mixed data types, providing transparency as each data characteristic and its role in clustering are explicitly defined. This stands in contrast to approaches requiring Dimensionality Reduction, enhancing interpretability.
    \item \textbf{Threshold Management}: Transparent threshold management ensures clarity in boundary decisions within the clustering process. This transparency aids in understanding why specific data points are considered similar or dissimilar, enhancing the overall interpretability of the results.
\end{itemize}

\section{Results and Discussions}

\subsection{Evaluation and comparison of clusters}

This section begins by showcasing the functionality of our algorithm through small-scale examples. Subsequently, we delve into an analysis of the algorithm's outcomes when applied to a comprehensive, publicly available mixed dataset. Additionally, we present the results obtained using various state-of-the-art algorithms that we introduced. Finally, we engage in a discussion of the findings and propose potential enhancements. Supplementary results can be accessed on Github\footnote{The link will be provided after the review (double-blind)}.

\begin{figure}[ht]
  \centering
  \includegraphics[width=\columnwidth]{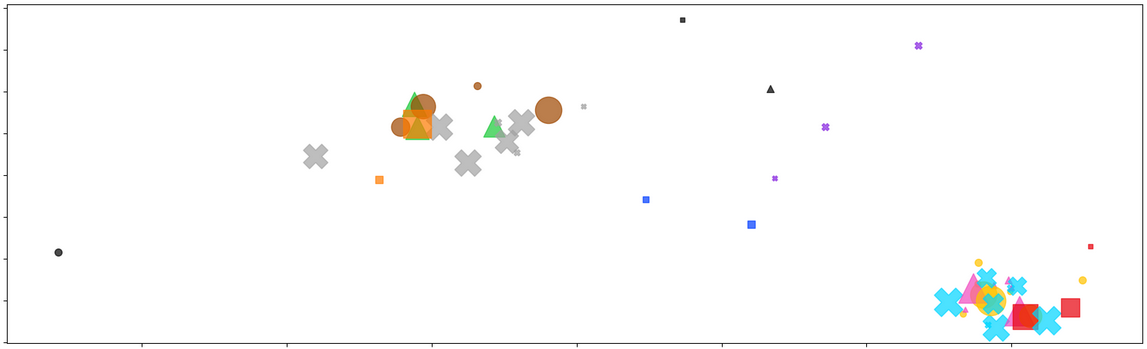}
  \caption{Clusters identified by our algorithm by taking into account the position, size and shape using the DNF of Position \textbf{and} size \textbf{and} shape.}
  \label{fig:p&s&s}
\end{figure}

In Figure \ref{fig:p&s&s}, we present a generated dataset comprising elements characterized by three attributes: their positions in a two-dimensional space, their size, and their shape. For each attribute, a weighted graph, termed a prenetwork, is created, where node weights indicate the similarity between elements based on that specific attribute. In this instance, Euclidean distance computes the similarity in the 2D space, absolute differences quantify size similarity, and Hamming distance evaluates shape similarity. Various calculation methods for these attributes can be applied, as the chosen method is a parameter of the prenetworks. Subsequently, the pretopological space is defined by establishing the DNF that links the prenetworks.

\begin{figure*}[ht]
  \centering
  \includegraphics[width=\textwidth]{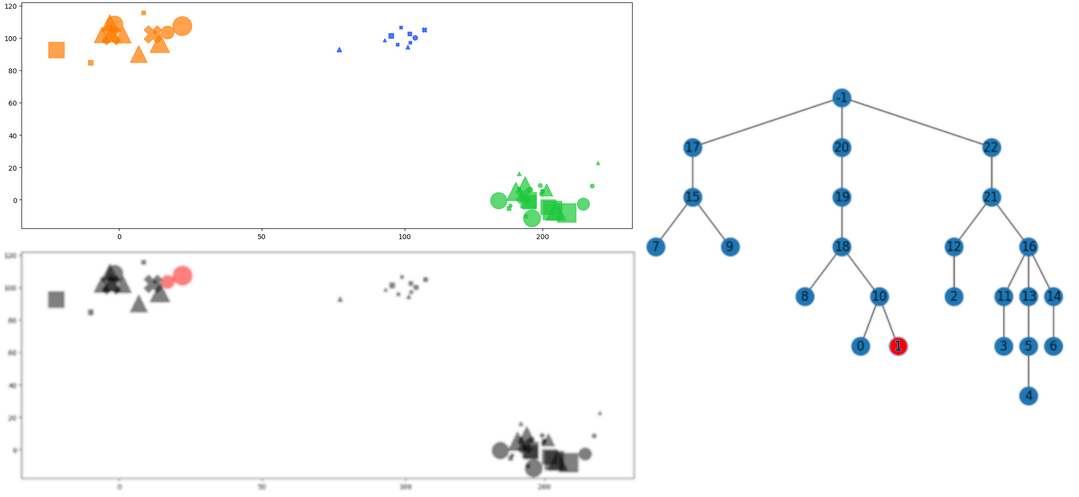}
  \caption{Clusters identified by our algorithm by taking into account the position, size and shape using the DNF of (Position \textbf{and} size) \textbf{or} (Position \textbf{and} shape). The subcluster "1" and its position in the hierarchy are colored in red.}
  \label{fig:p&s||p&s}
\end{figure*}

\begin{figure}[ht]
   \centering
   \includegraphics[width=\columnwidth]{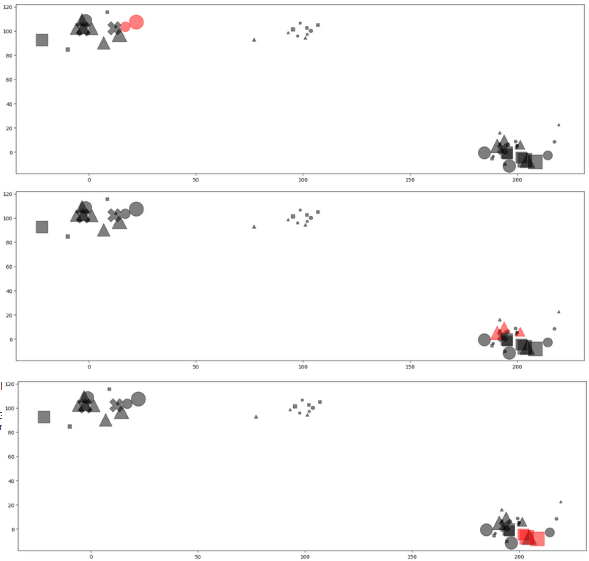}
   \caption{Different subsets of the clusters identified in Figure \ref{fig:p&s||p&s} are identified, we can see how the DNF has influenced the way the clusters are defined.}
   \label{fig:subsets}
\end{figure}

Figure \ref{fig:p&s&s} visually represents clustering results using the DNF: Position AND Size AND Shape. This DNF generates multiple clusters of elements that exhibit similarity across all attributes. In Figure \ref{fig:p&s||p&s}, we showcase the clustering outcome on another generated dataset using the DNF: (Position AND Size) OR (Position AND Shape). This DNF forms three large clusters of closely positioned elements in the 2D space. An analysis of the hierarchy of subclusters identified by our algorithm reveals that these subclusters consist of elements closely located in the 2D space and sharing similar shapes and/or sizes (Figure \ref{fig:subsets}). This underscores that by defining a specific pretopological prenetwork, a hierarchical structure of clusters tailored to specific requirements can be established. In this context, the three final clusters are obtained by selecting the sets with "-1" as a parent in the dendrogram (i.e., sets with no parent or whose parent is the entire set of elements).

\subsection{Evaluation of the clustering methods}

The assessment of various clustering algorithms is conducted utilizing the clustering quality measures outlined in the article. The Calinski-Harabasz, Davies-Bouldin, and Silhouette scores are computed by transforming datasets into Euclidean spaces through FAMD. The output space maintains the same dimensionality as the initial space, and the selection of FAMD is based on the following considerations:

\begin{itemize}
    \item It is a factorial method, the inertia of the model is known.
    \item It is deterministic.
    \item It does not rely heavily on hyper-parameters.
\end{itemize}

The Silhouette score, unique in its ability to accept a pairwise distance matrix as input, is computed using the Gower matrix. This approach is adopted to mitigate potential bias towards FAMD and to contribute to the analysis, particularly in cases where FAMD yields low inertia.

It is noteworthy that certain algorithms may yield either a single cluster or solely identify outliers. In such instances, the results table indicates "-" to signify these scenarios.

\subsection{Palmer Penguins}

The Palmer Penguins dataset, compiled from physical measurements of 344 penguins in the Palmer Archipelago, Antarctica \cite{gorman2014ecological}, serves as a foundational case study. Featuring four numerical and four categorical features, it is a commonly employed dataset in literature due to its typical structure and pronounced clustering tendencies across various dimension reductions.

Several clustering algorithms, such as DenseClus and those leveraging the Elbow Method, consistently partition the Palmer Penguins dataset into two clusters. The Elbow Method, by design, dictates a fixed number of clusters—in this case, two—resulting in identical outcomes for algorithms adhering to its constraints. The PretopoMD algorithm also identifies two clusters, albeit with a more equitable distribution; however, these clusters yield lower scores in the chosen evaluation metrics.

Contrastingly, the pretopological algorithm, implemented in three iterations with dimensionality reduction, presents diverse outcomes. The UMAP variant identifies three distinct clusters, PaCMAP discerns eleven clusters along with 112 outliers, and the FAMD iteration subdivides the data into twenty-six clusters. Intriguingly, the FAMD iteration, which produces the most clusters, showcases superior performance indices, suggesting its delineation captures a rich set of intrinsic dataset information. Notably, the FAMD inertia for this dataset peaks at an impressive 98.2\%, indicating a high representation of data variance. Refer to Table \ref{measures-penguins} for specific evaluation metric scores.

\begin{table}[]
\centering
\begin{tabular}{|l|c|c|c|c|}

\hline
 & \begin{tabular}[c]{@{}l@{}}Calinski\\ Harabasz\end{tabular} & \begin{tabular}[c]{@{}l@{}}Silhouette\\ FAMD\end{tabular} & \begin{tabular}[c]{@{}l@{}}Silhouette\\ Gower\end{tabular} & \begin{tabular}[c]{@{}l@{}}Davies\\ Bouldin\end{tabular} \\\hline\hline
K-Prototypes                                                 & 176.08                                                      & 0.34                                                      & 0.44                                                       & 1.23                                                     \\\hline
\begin{tabular}[c]{@{}l@{}}Modha\\ Spangler\end{tabular}     & 176.08                                                      & 0.34                                                      & 0.44                                                       & 1.23                                                     \\\hline
KAMILA                                                       & 176.08                                                      & 0.34                                                      & 0.44                                                       & 1.23                                                     \\\hline
ClustMD                                                      & 176.08                                                      & 0.34                                                      & 0.44                                                       & 1.23                                                     \\\hline
MixtComp                                                     & 176.08                                                      & 0.34                                                      & 0.44                                                       & 1.23                                                     \\\hline
\begin{tabular}[c]{@{}l@{}}Phillip \&\\ Ottaway\end{tabular} & 176.08                                                      & 0.34                                                      & 0.44                                                       & 1.23                                                     \\\hline
DenseClus                                                   & 176.08                                                      & 0.34                                                      & 0.44                                                       & 1.23                                                     \\\hline
\begin{tabular}[c]{@{}l@{}}Pretopo\\ FAMD\end{tabular}       & \textbf{182.22}                                             & \textbf{0.66}                                             & \textbf{0.65}                                              & \textbf{0.70}                                            \\\hline
\begin{tabular}[c]{@{}l@{}}Pretopo\\ UMAP\end{tabular}       & 163.89                                                      & 0.36                                                      & 0.44                                                       & 1.24                                                     \\\hline
\begin{tabular}[c]{@{}l@{}}Pretopo\\ PaCMAP\end{tabular}     & 61.80                                                       & 0.34                                                      & 0.29                                                       & 1.12           \\  \hline   
PretopoMD         &         105.17 &         0.239 &          0.263 &        1.71 \\ \hline
\end{tabular}
\caption{Results of the selected Algorithms on the Palmer Penguins dataset.}
\label{measures-penguins}
\end{table}

\subsection{Sponge}

The Sponge dataset, characterized by numerous categorical features and a relatively modest sample size, presents distinct characteristics when subjected to clustering algorithms. When applying FAMD to this dataset, a lower inertia of 86.13\% is observed compared to the Penguin dataset. The Hopkins statistic, indicative of the dataset's clustering tendency, is low at 0.63, as supported by the iVAT visualization, which reveals no clear evidence of inherent cluster structures.

In contrast, PaCMAP yields a significantly higher Hopkins statistic (0.88) and a more discernible iVAT, implying a comparatively simpler task for subsequent clustering algorithms, though not necessarily leading to superior clusters.

The evaluation metrics for the Sponge dataset, including Calinski-Harabasz scores, are approximately an order of magnitude lower than those observed for the Penguin dataset, indicating the presence of less well-defined clusters. Notably, K-prototype and KAMILA achieve the highest CH scores, followed by Pretopo-PaCMAP and Modha-Spangler. The Silhouette Gower score is highest for Phillip \& Ottaway.

Pretopological FAMD clustering exhibits a notably low Davies-Bouldin score, identifying 63 outliers and forming 6 clusters, each consisting of 2 to 3 elements. While the clusters are distinct, the other indices are significantly low, emphasizing that the clusters do not demarcate a clear partition of the dataset. This highlights the importance of using multiple indicators to gauge clustering quality.

Conversely, PretopoMD identifies a single, dominant cluster comprising 74 elements and only one outlier. Given the dataset's weak clustering propensity, this outcome seems relevant, with PretopoMD boasting the highest FAMD Silhouette score and the lowest DB score. For specific metric scores, refer to the detailed results in the respective evaluation Table \ref{measures-sponge}.

\begin{table}[]
\centering
\begin{tabular}{|l|c|c|c|c|}
\hline
 & \begin{tabular}[c]{@{}l@{}}Calinski\\ Harabasz\end{tabular} & \begin{tabular}[c]{@{}l@{}}Silhouette\\ FAMD\end{tabular} & \begin{tabular}[c]{@{}l@{}}Silhouette\\ Gower\end{tabular} & \begin{tabular}[c]{@{}l@{}}Davies\\ Bouldin\end{tabular} \\\hline\hline
K-Prototypes       & \textbf{16.83}    & 0.170  & 0.421            & 2.01           \\ \hline
\begin{tabular}[c]{@{}l@{}}Modha\\ Spangler\end{tabular}     & 16.72             & 0.168           & 0.404            & 2.03           \\ \hline
KAMILA             & \textbf{16.83}    & 0.170  & 0.421            & 2.01           \\ \hline
ClustMD            & -                 & -               & -                & -              \\ \hline
MixtComp           & -                 & -               & -                & -              \\ \hline
\begin{tabular}[c]{@{}l@{}}Phillip \&\\ Ottaway\end{tabular} & 15.66             & 0.161           & \textbf{0.437}   & 2.09           \\ \hline
DenseClus          & -                 & -               & -                & -              \\ \hline
\begin{tabular}[c]{@{}l@{}}Pretopo\\ FAMD\end{tabular}       & 1.65              & -0.061          & -0.168           & 1.46  \\ \hline
\begin{tabular}[c]{@{}l@{}}Pretopo\\ UMAP\end{tabular}       & 7.53              & 0.084           & 0.142            & 2.79           \\ \hline
\begin{tabular}[c]{@{}l@{}}Pretopo\\ PaCMAP\end{tabular}     & 16.76             & 0.169           & 0.431            & 2.00           \\ \hline
PretopoMD         &           6.36 &         \textbf{0.484} &          0.012 &        \textbf{0.383} \\ \hline
\end{tabular}
\caption{Results of the selected Algorithms on the Sponge dataset.}
\label{measures-sponge}
\end{table}

\subsection{Dataset generator}

To evaluate the different algorithms over every desired configurations, we use a dataset generator. The most common way to generate datasets to benchmark and evaluate clustering algorithms is to generate isotropic gaussian blobs. This method is  natively present in the widely used scikit-learn for Python by \cite{scikit-learn}, MixSim for R  by \cite{melnykov2012mixsim} and Linfa for Rust\footnote{\url{https://rust-ml.github.io/linfa/}}. 

First, we generated cluster centers, with an average pairwise distance of 1. Then we generate samples from a gaussian mixture model with the density described by:

\begin{equation}
    p(x) = \frac{1}{k} \sum_{i=1}^{k} \mathcal{N}(\mu_i, \Sigma_i)
    \label{generator-density}
\end{equation}
where: \\
    - $k$ is the number of clusters;
    - $\mu_i$ are the cluster centers;
    - $\Sigma_i$ refers to the cluster covariances. Here, it is a diagonal matrix of the clusters variance.  
    
Inspired by \cite{costa2022benchmarking}, we split features upon quantiles to transform them into categorical features. Thus, we get a mixed dataset. With this method, the different parameters we can tune to obtain different configurations are:
\begin{itemize}
    \item The number of samples to generate (the number of individuals);
    \item The number of clusters $k$;
    \item The number of numerical features;
    \item The number of categorical features;
    \item The number of unique values taken by categorical variables;
    \item The standard deviation of clusters.
\end{itemize}

\subsection{Generated dataset with 3 clusters}

In the case of the Base Generated dataset with 500 individuals, 5 features (numerical/categorical), 3 clusters, 3 categorical unique values, and a standard deviation of 0.1, the Elbow Method suggests $k=3$ as the optimal number of clusters. This alignment with the intended number of clusters provides an advantage to algorithms utilizing the Elbow Method.

As a result, algorithms employing the Elbow Method produce very similar partitions of the dataset, with closely aligned results across all four evaluation indices. The two algorithms utilizing UMAP, DenseClus, and Pretopo-UMAP, also yield similar results even without relying on the Elbow Method.

On the other hand, Pretopo-FAMD reports $444$ outliers, and Pretopo-PaCMAP identifies six clusters with $124$ outliers. Meanwhile, PretopoMD detects three sizable clusters and $60$ outliers. As observed in other datasets, adjusting the hyperparameters for PretopoMD might lead to either improvements or deteriorations in the obtained results. For detailed metric scores, refer to the specific evaluation table showcasing the results for the Base Generated Case dataset (Table \ref{basecase}).

\begin{table}
\centering
\begin{tabular}{|l|c|c|c|c|} 
\hline

 & \begin{tabular}[c]{@{}l@{}}Calinski\\ Harabasz\end{tabular}  & \begin{tabular}[c]{@{}l@{}}Silhouette\\ FAMD\end{tabular} & 
 \begin{tabular}[c]{@{}l@{}}Silhouette\\ Gower\end{tabular} & 
 \begin{tabular}[c]{@{}l@{}}Davies\\ Bouldin\end{tabular} \\\hline\hline
 
ClustMD & \textbf{296.330} & \textbf{0.394} & \textbf{0.521} & \textbf{1.071} \\\hline
DenseClus & 295.224 & 0.393 & 0.519 & 1.074 \\\hline
Phillip \& Ottaway & 290.556 & 0.389 & 0.514 & 1.083 \\\hline
Kamila & \textbf{296.330} & \textbf{0.394} & \textbf{0.521} & \textbf{1.071} \\\hline
K-Prototypes & \textbf{296.330} & \textbf{0.394} & \textbf{0.521} & \textbf{1.071} \\\hline
MixtComp & 294.906 & 0.393 & 0.518 & 1.075 \\\hline
Modha-Spangler & \textbf{296.330} & \textbf{0.394} & \textbf{0.521} & \textbf{1.071} \\\hline
Pretopo-FAMD & 34.160 & 0.067 & 0.061 & 1.453 \\\hline
Pretopo-UMAP & 293.835 & 0.392 & 0.516 & 1.078 \\\hline
Pretopo-PaCMAP & 66.563 & 0.049 & 0.013 & 1.845 \\\hline
PretopoMD & 127.338 & 0.230 & 0.308 & 1.508 \\\hline
\end{tabular}
\caption{Results of the selected Algorithms on the Base Generated Case.}
\label{basecase}
\end{table}

\subsection{Generated Dataset with 10 clusters} 
For the generated dataset with 10 clusters, created with the same parameters as the previous case but with the distinction of having 10 clusters, the Elbow Method deviates from the intended number of clusters, identifying $k=2$ as optimal.

Pretopo FAMD identifies $470$ outliers and pinpoints $14$ small clusters, each containing fewer than $10$ individuals. Despite this seemingly suboptimal clustering, it achieves the best Davies-Bouldin score. DenseClus and PretopoMD both identify approximately $300$ outliers among the $500$ data points, even though the generated datasets are not designed to contain noise.

Lastly, Pretopo-UMAP divides the dataset into $8$ distinct clusters and achieves the highest scores across both Silhouette versions and the Calinski-Harabasz index. It also holds the second-best position for the Davies-Bouldin score. For detailed metric scores, refer to the specific evaluation Table \ref{measures} showcasing the results for the Generated Dataset with $10$ Clusters.

\begin{table}[]
\centering
\begin{tabular}{|l|c|c|c|c|}
\hline
 & \begin{tabular}[c]{@{}l@{}}Calinski\\ Harabasz\end{tabular} & \begin{tabular}[c]{@{}l@{}}Silhouette\\ FAMD\end{tabular} & \begin{tabular}[c]{@{}l@{}}Silhouette\\ Gower\end{tabular} & \begin{tabular}[c]{@{}l@{}}Davies\\ Bouldin\end{tabular} \\\hline\hline
ClustMD & 95.934 & 0.175 & 0.197 & 2.216 \\ \hline
DenseClus & 81.313 & 0.183 & 0.200 & 1.914 \\ \hline
Phillip \& Ottaway & 70.949 & 0.194 & 0.213 & 1.190 \\ \hline
Kamila & 86.348 & 0.160 & 0.157 & 2.120 \\ \hline
K-Prototypes & 92.096 & 0.168 & 0.185 & 2.263 \\ \hline
MixtComp & 83.316 & 0.163 & 0.205 & 2.163 \\ \hline
Modha-Spangler & 79.214 & 0.153 & 0.206 & 2.421 \\ \hline
Pretopo-FAMD & 14.482 & -0.015 & -0.018 & 2.219 \\ \hline
Pretopo-UMAP & \textbf{127.624} & \textbf{0.353} & \textbf{0.359} & \textbf{1.122} \\ \hline
Pretopo-PaCMAP & 66.870 & 0.174 & 0.154 & 1.557 \\ \hline
PretopoMD & 48.178 & 0.108 & 0.164 & 2.197 \\ \hline
\end{tabular}
\caption{Results of the selected Algorithms on a generated dataset with 10 clusters.}
\label{measures}
\end{table}

\subsection{Generated dataset of high dimension}

Then, we analyze how the different algorithms perform in a high dimension context (15 quantitative features and 15 categorical features). To do so, we generate a dataset with 15 clusters of each size. 

There, the Elbow Method finds $k = 2$ clusters. ClustMD, PretopoMD, pretopo UMAP and MixtComp don't converge on such a dataset, and only produce noise. Pretopo FAMD finds $498$ outliers out of the $500$ individuals. Pretopo UMAP produces $1$ cluster of $332$ individuals and $168$ outliers (that might be merged into another cluster). DenseClus and Pretopo PaCMAP both find $3$ balanced clusters, and the latter obtains a slightly better score on the $4$ indices. 

Kamila, K-prototype and Phillip and Ottaway have merged two clusters into one and therefore find a cluster of approximatively $333$ elements and another one of around $137$ elements. For a detailed breakdown of metric scores, please refer to the dedicated evaluation Table \ref{measures-15cat-15catuniques}.

\begin{table}
\centering
\begin{tabular}{|l|c|c|c|c|} 
\hline

 & \begin{tabular}[c]{@{}l@{}}Calinski\\ Harabasz\end{tabular}  & \begin{tabular}[c]{@{}l@{}}Silhouette\\ FAMD\end{tabular} & 
 \begin{tabular}[c]{@{}l@{}}Silhouette\\ Gower\end{tabular} & 
 \begin{tabular}[c]{@{}l@{}}Davies\\ Bouldin\end{tabular} \\\hline\hline
 
ClustMD & 0.000 & -1.000 & -1.000 & -1.000 \\ \hline
DenseClus & 118.048 & 0.218 & 0.102 & 1.723 \\ \hline
Phillip \& Ottaway & 119.284 & 0.195 & 0.087 & 1.831 \\ \hline
Kamila & 120.934 & 0.196 & 0.088 & 1.817 \\ \hline
K-Prototypes & 118.048 & 0.191 & 0.085 & 1.862 \\ \hline
MixtComp & 0.000 & -1.000 & -1.000 & -1.000 \\ \hline
Modha-Spangler & 120.934 & 0.196 & 0.088 & 1.817 \\ \hline
Pretopo-FAMD & 0.631 & -0.030 & -0.001 & 3.079 \\ \hline
Pretopo-UMAP & 0.000 & -1.000 & -1.000 & -1.000 \\ \hline
Pretopo-PaCMAP & \textbf{122.833} & \textbf{0.227} & \textbf{0.107} & \textbf{1.681} \\ \hline
PretopoMD & 0.000 & -1.000 & -1.000 & -1.000 \\ \hline
\end{tabular}
\caption{Results of the selected Algorithms on a generated dataset with 15 quantitative features and 15 categorical features.}
\label{measures-15cat-15catuniques}
\end{table}

\subsection{Generated Dataset with sparse clusters}

The performance of various algorithms was studied on datasets with sparser clusters, specifically a deviation of 0.15, in contrast to the base case's deviation of 0.10. The dataset comprises 1000 individuals, with 10 dimensions of each type. The Elbow method suggests $k=2$ as the optimal number of clusters, while the dataset is intended to contain 3 clusters.

The algorithms that obtain the optimal scores are the algorithms that use UMAP and PaCMAP. As those reduction move the neighbors closer to each other, it is not surprising to see them perform well on datasets with a higher clusters deviation. Among those 3, Pretopo PaCMAP is the only one that produces no outlier, therefore it obtains the band Silhouette scores. For a detailed breakdown of metric scores, please refer to the dedicated evaluation Table \ref{measures-harder-dataset}.

\begin{table}[]
\centering
\begin{tabular}{|l|c|c|c|c|}
\hline
 & \begin{tabular}[c]{@{}l@{}}Calinski\\ Harabasz\end{tabular} & \begin{tabular}[c]{@{}l@{}}Silhouette\\ FAMD\end{tabular} & \begin{tabular}[c]{@{}l@{}}Silhouette\\ Gower\end{tabular} & \begin{tabular}[c]{@{}l@{}}Davies\\ Bouldin\end{tabular} \\\hline\hline
ClustMD & 184.929 & 0.150 & 0.132 & 2.109 \\ 
\hline
DenseClus & 133.900 & 0.144 & 0.142 & 2.987 \\ 
\hline
Phillip \& Ottaway & 5.673 & 0.100 & 0.157 & 2.558 \\ 
\hline
Kamila & 184.963 & 0.150 & 0.132 & 2.117 \\ 
\hline
K-Prototypes & 181.246 & 0.148 & 0.128 & 2.145 \\ 
\hline
MixtComp & 169.897 & 0.142 & 0.144 & 2.204 \\ 
\hline
Modha-Spangler & 169.793 & 0.143 & 0.145 & 2.202 \\ 
\hline
Pretopo-FAMD & 1.199 & 0.017 & 0.026 & 2.160 \\ 
\hline
Pretopo-UMAP & 162.758 & 0.136 & 0.135 & 2.261 \\ 
\hline
Pretopo-PaCMAP & \textbf{202.114} & \textbf{0.191} & \textbf{0.190} & \textbf{1.848} \\ 
\hline
PretopoMD & 17.020 & -0.032 & -0.055 & 2.800 \\ 
\hline
\end{tabular}
\caption{Results of the selected Algorithms on a generated dataset with with 1000 individuals, 10 dimensions of each type, and a deviation of 0.15}
\label{measures-harder-dataset}
\end{table}

\subsection{Measures and comparison of performances}

To assess the performance of various clustering methods, we conducted experiments on several generated datasets with diverse specifications. The execution statistics presented here were obtained on a system equipped with an AMD Ryzen 7 5800H CPU, operating at a frequency of 3.20GHz, featuring 512KB of L1 cache and 32GB of DDR4 RAM. The detailed results are accessible on GitHub, and a more comprehensive discussion will be provided in an upcoming paper.

\begin{figure*}[ht]
  \centering
  \includegraphics[width=\textwidth]{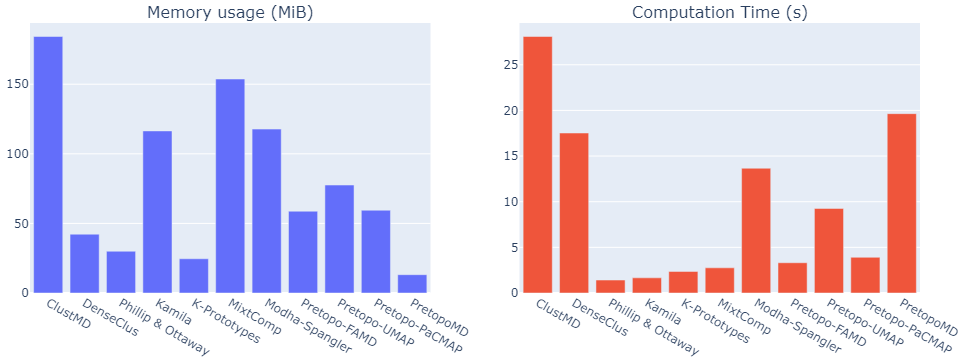}
  \caption{Memory usage and computation time on a $500$ elements generated dataset. With $5$ numerical features and $5$ categorical features.}
  \label{fig:memory&computation}
\end{figure*}

\begin{figure*}[ht]
  \centering
  \includegraphics[width=\textwidth]{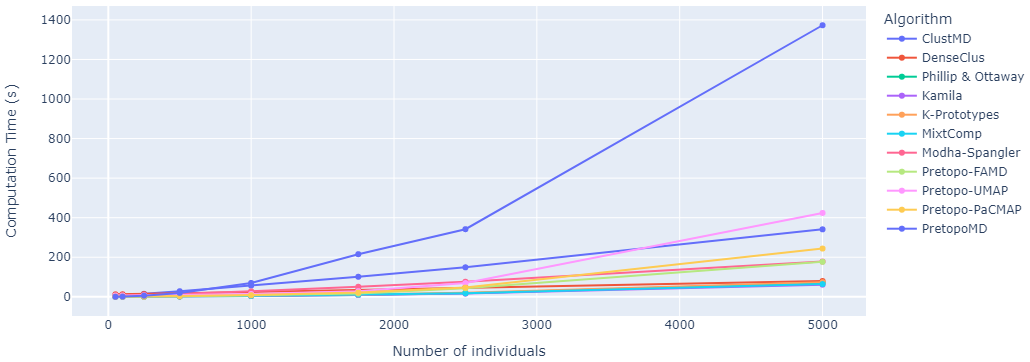}
  \caption{Computation time per number of elements.}
  \label{fig:computation_per_elements}
\end{figure*}

Memory Usage and Computational Time Comparison (Dataset: 500 Elements, 5 Numerical Features, 5 Categorical Features)
Figure \ref{fig:memory&computation} offers a visual representation comparing the memory usage and computational time of different clustering methods. Notable observations include:

\begin{itemize}
    \item PretopoMD Algorithm: Demonstrates the lowest memory usage but requires substantial computation time. Analyzing datasets of varying sizes indicates that, for this implementation of the algorithm, computation time scales linearly with the number of individuals and remains invariant with respect to the number of features (refer to Figure \ref{fig:computation_per_elements}).
    \item ClustMD: Exhibits the highest computation time among the algorithms considered.
\end{itemize}

\subsection{Comparison with SOTA methods}

\added{new subsection}

PretopoMD distinguishes itself by directly processing mixed data without requiring preliminary dimensionality reduction, thereby mitigating the risks of information loss and projection bias commonly encountered in high-dimensional transformations. Its capability to operate on heterogeneous data types ensures that the intrinsic variability of numerical and categorical features is preserved. Furthermore, PretopoMD yields a hierarchical clustering structure that produces an interpretable dendrogram, enabling the exploration of nested, multi-level clusters without the necessity to predefine the number of clusters. The integration of customizable logical rules further enhances the method's interpretability, allowing researchers to incorporate domain-specific knowledge into the clustering process and derive more meaningful insights.

In contrast, methods such as K-Prototypes and Convex K-Means, although computationally efficient, require pre-specification of cluster counts. For instance, K-Prototypes combines Euclidean and matching distances to jointly handle numerical and categorical data but relies on a simplistic categorical distance measure and necessitates careful tuning of balancing parameters, potentially limiting its effectiveness. Similarly, Convex K-Means improves convergence through convex-hull-based refinement of centroid positions; however, its additional computational overhead and the inherent requirement for a predetermined number of clusters restrict its applicability in more complex scenarios.

Model-based approaches, including MixtComp, Kamila, and ClustMD, offer a probabilistic framework through latent variable modeling that can provide deep insights into the underlying data structure. Despite their rigorous statistical foundations and potential for rich interpretability, these methods are often hampered by high computational costs and convergence issues related to Bayesian estimation or parameter initialization. In applications demanding rapid and intuitive clustering outcomes, such complexities may render these approaches less practical.

Overall, PretopoMD's strengths lie in its flexibility and enhanced explainability. By eliminating the need for a preset number of clusters and delivering a dendrogram that integrates logical rules, it fosters a transparent clustering process especially advantageous for domains where interpretability is paramount. However, this benefit is counterbalanced by its significant computational complexity—stemming from the iterative pseudoclosure procedure and adjacency matrix construction—which becomes more pronounced when dealing with large or high-dimensional datasets. Moreover, the method's performance is highly sensitive to hyperparameter settings, necessitating meticulous tuning and substantial domain expertise.

In summary, while PretopoMD provides a robust and explainable framework for clustering mixed data, its adoption requires careful consideration of the trade-offs between computational efficiency and interpretability. In environments where scalability and rapid computation are essential, simpler techniques like K-Prototypes or Convex K-Means may be preferable. Conversely, in scenarios that prioritize the interpretability of clusters and the ability to capture subtle, domain-specific patterns, PretopoMD represents a compelling alternative despite its higher computational demands.

\subsection{Discussion}

Experimental results indicate that clustering performance often improves when applied after dimensionality reduction. This improvement can be partially attributed to the Curse of Dimensionality—a phenomenon that arises in high-dimensional spaces, particularly in clustering and machine learning tasks, where the exponential growth of search spaces hinders algorithm efficiency \cite{beyer1999nearest, verleysen2005curse}. Moreover, distance metrics that perform well in lower-dimensional spaces may lose their effectiveness in higher dimensions \cite{steinbach2004challenges}, making dimensionality reduction a valuable tool to alleviate these issues.

A detailed analysis of clustering outcomes using PretopoMD, in comparison with other methods across diverse datasets, reveals that the algorithm is sensitive to dataset characteristics such as dimensionality, noise level, and cluster dispersion. For example, PretopoMD demonstrates strong performance on balanced datasets with well-defined cluster boundaries—evidenced by competitive evaluation metrics on the Base Generated dataset—but shows reduced performance on high-dimensional or noisy datasets, as observed with the Sponge dataset. This variability is largely attributable to the algorithm’s dependency on carefully tuned hyperparameters, including threshold values, logical rules embedded in DNF, and seed sizes.

A comprehensive sensitivity analysis is underway to quantify the impact of these hyperparameters on cluster quality. Preliminary results indicate that even subtle variations in key parameters can result in differences in both the number of detected clusters and the proportion of outliers. Additionally, the hierarchical structure of PretopoMD enables the construction of dendrograms, which offer intuitive visualizations of cluster relationships.

These findings underscore the critical need for hyperparameter tuning and dataset-specific customization when applying PretopoMD. Although the algorithm offers significant advantages—such as handling mixed data without prior dimensionality reduction and enhancing explainability through customizable parameters—its optimal performance relies on human expertise. Elements such as threshold values, seed selection, and logical rules require careful adjustment to balance computational efficiency with clustering quality, particularly in high-dimensional or noisy contexts.

Additionally, we are developing a DNF-based clustering approach in collaboration with energy experts. This approach aims to cluster buildings based on features like construction material and construction date, which are presumed to correlate with latent characteristics affecting energy consumption. The use of DNF enables us to explicitly formulate logical rules (e.g., "feature A above a certain threshold OR feature B within a specific range"), thereby improving interpretability. However, it is important to acknowledge that while DNF promotes clarity, its simplicity may restrict expressiveness in datasets with complex interdependencies, potentially reducing cluster granularity.

To balance explainability with clustering quality, several strategies can be employed:
\begin{itemize}
    \item \textbf{Multi-Stage Clustering Approach:} Initially conduct a fine-grained clustering that emphasizes the fidelity of the data. Subsequently, abstract the results into interpretable logical rules, preserving subtle patterns while ultimately presenting a clear summary of the clusters.
    \item \textbf{Automated Hyperparameter Tuning:} Utilize optimization techniques combined with internal (e.g., Silhouette score, Calinski-Harabasz index) and external metrics (e.g., ARI, MI) to determine the most appropriate hyperparameter settings. This approach helps in achieving an optimal balance between clustering fidelity and explainability.
    \item \textbf{Iterative Expert Feedback:} Engage domain experts in iterative evaluation cycles. Expert insights can guide further tuning of hyperparameters and the refinement of logical rules, ensuring that the clustering outcomes are both high in quality and consistent with domain-specific expectations.
\end{itemize}

By adopting these strategies, it becomes possible to harness the strengths of PretopoMD—namely its performance and explainability—while mitigating trade-offs between clarity and cluster granularity. This balanced approach is crucial, as high scores on quality indicators such as the Calinski-Harabasz index, Silhouette coefficient (computed with FAMD or Gower distances), or the Davies-Bouldin index do not necessarily coincide with the expectations of domain experts. In many real-world applications, experts may prioritize clusters that resonate with domain-specific insights—for instance, grouping buildings with similar operational constraints—even if such groupings yield comparatively lower scores on generic quality metrics. This observation highlights the necessity for human-centered evaluation criteria, underscoring that clustering should be approached not only as an optimization problem but also as a tool for deriving meaningful, domain-relevant insights.

\section{Conclusion}
This study addresses the intricate challenges of clustering mixed data, with particular emphasis on applications in the energy sector and other diverse fields. We provide a comprehensive overview of existing clustering methods and dimensionality reduction techniques for mixed data, critically discussing their strengths and limitations. A central theoretical contribution of this work is the introduction of the pretopology-based algorithm, pretopoMD, which offers a novel perspective on how mixed data can be effectively clustered.

PretopoMD presents several key innovations. It allows for Customizable Logical Rules, enabling the definition of tailored rules for cluster construction that provide flexibility to meet specific application requirements. It also features Adjustable Hyperparameters, which facilitate fine-tuning for clustering and division conditions and thus offer precise control over the clustering process. Additionally, the algorithm supports Explainable Cluster Construction by enabling hierarchical dendrogram analysis that enhances the transparency and interpretability of clustering outcomes.

Our experimental results underscore that, although pretopoMD may involve relatively higher computational times, it exhibits low memory usage—a crucial advantage when handling large-scale datasets. While these outcomes validate its potential as a robust and interpretative solution for mixed data clustering, the study also acknowledges limitations, such as the increased computational burden and the need for further hyperparameter optimization. Future research should therefore focus on improving the algorithm's efficiency, exploring its applicability to richer energy datasets, and extending its use to other domains including biology, medicine, marketing, and economics.

In summary, this work contributes new theoretical insights and practical tools to the domain of mixed data clustering by integrating customization, fine-tuning, and transparency into its framework. These advances not only deepen our understanding of clustering heterogeneous data but also pave the way for future innovations in managing complex datasets across various fields.

\backmatter


\section*{Declarations}

\subsection*{Funding}

This paper is the result of research conducted at the energy data management company \textit{Energisme}. We thank \textit{Energisme} for the resources that have been made available to us and Julio Laborde for his assistance with the conception of our pretopological hierarchical algorithm library.

\subsection*{Conflict of interest/Competing interests}

Not applicable.

\subsection*{Ethics approval and consent to participate}

Not applicable.

\subsection*{Consent for publication}

All authors consent for publication.

\subsection*{Data availability}

All datasets are given in the github \url{https://github.com/ClementCornet/Benchmark-Mixed-Clustering}.

\subsection*{Materials availability}

All methods with a README are given in the github \url{https://github.com/ClementCornet/Benchmark-Mixed-Clustering}.

\subsection*{Code availability}

All codes are given in the github \url{https://github.com/ClementCornet/Benchmark-Mixed-Clustering}.

\subsection*{Author contribution}

Authors equally contribute to this paper.

\bibliography{biblio}

\end{document}